\pdfoutput=1
\documentclass[11pt]{article}

\usepackage[preprint]{acl}

\usepackage{times}
\usepackage{latexsym}

\usepackage[T1]{fontenc}
\usepackage[utf8]{inputenc}

\usepackage{microtype}

\usepackage{inconsolata}

\usepackage{graphicx}
\usepackage{amsmath}
\usepackage{multirow}
\usepackage{booktabs}
\usepackage{colortbl}
\usepackage{color}
\usepackage{amssymb}
\usepackage{diagbox}

\usepackage{tcolorbox}
\tcbuselibrary{most}

\newtcolorbox{promptbox}{
    colback=gray!5,
    colframe=gray!20,
    fonttitle=\bfseries,
    breakable,  % 支持跨页
}

\title{InfiniteICL: Breaking the Limit of Context Window Size \\ via Long Short-term Memory Transformation}

\author{Bowen Cao$^{{\diamondsuit}}$ \quad Deng Cai \quad Wai Lam$^{\diamondsuit}$ \\
$\diamondsuit$ The Chinese University of Hong Kong \ \\
\texttt{bwcao@link.cuhk.edu.hk}, \ \texttt{thisisjcykcd@gmail.com}, \ \texttt{wlam@se.cuhk.edu.hk} \\
}

\begin{document}
\maketitle
\begin{abstract}
In-context learning (ICL) is critical for large language models (LLMs), but its effectiveness is constrained by finite context windows, particularly in ultra-long contexts. To overcome this, we introduce \textbf{InfiniteICL}, a framework that parallels context and parameters in LLMs with short- and long-term memory in human cognitive systems, focusing on transforming temporary context knowledge into permanent parameter updates. This approach significantly reduces memory usage, maintains robust performance across varying input lengths, and theoretically enables infinite context integration through the principles of context knowledge elicitation, selection, and consolidation. Evaluations demonstrate that our method reduces context length by $90\%$ while achieving $103\%$ average performance of full-context prompting across fact recall, grounded reasoning, and skill acquisition tasks. When conducting sequential multi-turn transformations on complex, real-world contexts (with length up to $2$M tokens), our approach surpasses full-context prompting while using only $0.4\%$ of the original contexts. These findings highlight InfiniteICL's potential to enhance the scalability and efficiency of LLMs by breaking the limitations of conventional context window sizes.
\end{abstract}

\section{Introduction}
% LLMs are powerful due to their remarkable ability to perform in-context learning, where the model is only provided with the learning materials in its input. Numerous studies have shown that LLMs can learn to complete diverse tasks with only an abstract task description and a few concrete demonstrations. This learning paradigm is unique to LLMs. The learning is not guided by explicit manually-designed training objectives and does not require gradient computing. In theory, it can utilize all kinds of data, including but not limited to demonstration data in supervised fine-tuning and preference data in RLHF.
In-context learning (ICL) has emerged as a cornerstone capability of large language models (LLMs), allowing training-free and user-friendly customization \cite{liu2024deepseek,yang2024qwen2,team2024gemma}. This ability is critical for real-world deployment in scenarios such as deep research integrating web information and lifelong learning \cite{zheng2025lifelong} through user interactions.

% Though the underlying mechanisms of ICL are still unclear, we find ICL might be the ultimate learning paradigm for LLMs for its flexibility and analogy to human learning. As humans, our experience (or memory) shapes our current mental state (intelligence). If an LLM can learn effectively from infinitely long contexts (e.g., the whole interaction history with a user), not only does this support a broad range of applications natively (e.g., long text generation, continuous knowledge update, and personalization), but also fits the ultimate goal of AGI. However, in-context learning is limited by the maximum context window.
However, the effectiveness of ICL is constrained by the finite context windows of the Transformer architecture \cite{waswani2017attention}, typically $8$K-$128$K tokens. This limitation manifests through two compounding factors: (\textit{i}) the quadratic complexity scaling of attention mechanism and the linear growth of KV cache memory \cite{liu2024deepseek}, which impose prohibitive hardware infrastructure requirements for long-context deployment, and (\textit{ii}) diminishing performance returns observed in tasks requiring extended contexts, such as many-shot learning \cite{agarwal2024many} and cross-document reasoning \cite{bai2024longbench}. These constraints create a paradox: expanding context capacity inflates computational costs disproportionately while delivering marginal accuracy gains.

\begin{figure}[t]
 \centering
 \includegraphics[width=\linewidth]{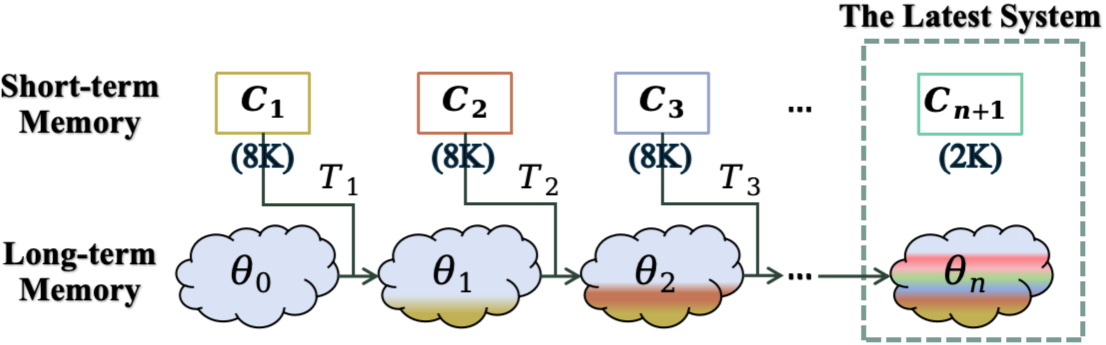}
 \caption{The core idea of our framework. The context window is refreshed after each transformation ($T_i: \theta_{i-1}+C_i\rightarrow \theta_i$), allowing infinite context input in a streaming fashion.
}
\label{fig:intro}
\end{figure}

In response to these challenges, we propose a novel perspective that frames the dichotomy between context and parameters in LLMs as analogous to short-term and long-term memory in human cognitive systems \cite{cowan2008differences}. Specifically, we posit that contexts function as short-term memory, capturing transient information relevant to the current input, while parameters serve as long-term memory, encoding accumulated knowledge over time. Building on this analogy, our approach focuses on transforming temporary context knowledge into permanent parameter updates, as depicted in Figure \ref{fig:intro}. This transformation draws inspiration from the human circadian rhythm, where daily experiences are consolidated into long-term memories during sleep \cite{klinzing2019mechanisms}. It is further supported by recent studies demonstrating that ICL can be interpreted as meta-gradient updates \cite{dai2023can,von2023transformers}. 
Additionally, our method falls within the scope of test-time compute scaling \cite{wei2022chain,wang2022self,akyurek2024surprising}, which enhances model capabilities through strategic allocation of compute for training during inference.

Compared to conventional strategies of extending context windows, our approach offers three key advantages: (\textit{i}) it reduces GPU memory usage by converting context into compact parameter updates; (\textit{ii}) it potentially maintains robust performance across varied input lengths, mitigating the scaling issues of attention mechanisms; and (\textit{iii}) it theoretically enables infinite context integration by continuously updating long-term memory, eliminating the need for ever-expanding context retention.

To implement this cognitive-inspired paradigm, we develop a framework with three principled design choices:  
(\textit{i}) Context knowledge elicitation establishes a unified strategy for diverse scenarios through hybrid prompting - systematically steering the model to generate both task-specific interactions (\textit{e.g.}, summarization, multi-step reasoning chains) and open-ended contextual expansions. This process constructs a transfer set $\mathcal{T}$ that comprehensively encodes contextual knowledge.
(\textit{ii}) Path selection optimizes $\mathcal{T}$ into $\mathcal{T}_k$ by retaining top-$k$ interactions with maximum perplexity discrepancy between context-aware and context-free generations, prioritizing knowledge-critical pathways.
(\textit{iii}) Memory consolidation transforms these temporary contextual insights into permanent model parameter updates through knowledge distillation.

Our framework is rigorously evaluated through a systematic protocol designed to comprehensively assess its effectiveness across diverse scenarios. We first examine our approach in the single transformation setting, where the model undergoes a one-time conversion of context into long-term memory. 
When transforming $90\%$ of the context into parameter updates, our method maintains $84\%$ performance for fact recall on Natural Questions, $117\%$ in grounded reasoning (Counterfact and Mquake), and $98\%$ in $300$-shot ICL (TREC and NLU), all compared to full-context prompting. These results collectively yield an average performance recovery of $103\%$ relative to full-context prompting, consistently outperforming state-of-the-art context compression methods, with average scores of $70.4$ versus $10.4$ for LLMLingua-2 \cite{pan2024llmlingua} and $7.3$ for SnapKV \cite{li2024snapkv}.

We further validate our framework in sequential transformation scenarios, where multi-turn parameter updates are conducted with consecutive contexts. On LongBench v2 \cite{bai2024longbench}, our method achieves superior results to full-context prompting (up to $128K$) while utilizing merely $0.4\%$ of the original contexts. Extended analysis reveals that existing context compression/distillation baselines exhibit progressive performance degradation as context lengths increase (e.g., relative performance declines range from $28\%$ to $82\%$ as context lengths grow from $8$K to $64$K tokens), with most baselines experiencing performance collapse at $2$M tokens. In contrast, our method maintains largely stable performance throughout this length scaling trajectory. Additionally, we carry out systematic ablations to study the choices and effects of the three components (context knowledge elicitation, selection, and consolidation) in our framework.

\section{Related Work}
\paragraph{Context Compression}
% (1) L1 -> L2
% (1.1) KV (1.2)token
% (1.1.1) group attention (1.1.2) quantization (1.1.3) on length (1.1.4) cross-layer  (1.1.5) channel 
% (1.2.1) delete tokens (1.2.2) summarize (1.2.3) GIST
% empirical comparsion with (1.2.1) (1.2.3) (1.1.3)

Some research focuses on reducing input context length to ensure it remains within a manageable computational budget. \citet{xu2023recomp,lee2024human}
employ summarization models to generate concise scripts. Other approaches remove redundant tokens based on information entropy \cite{li2023compressing,jiang2023llmlingua,jiang2023longllmlingua,pan2024llmlingua} or downstream task performance \cite{jung2024discrete,huang2023fewer}, the latter requiring task-specific training. Some works focus on learning soft task prompts that encodes contexts into trainable vectors \cite{wingate2022prompt,chevalier2023adapting,ge2023context,kim2023compressed,mu2024learning}. Additionally, there is a line of work that compresses context from the KV perspective, such as reducing KV activations along the length dimension \cite{li2024snapkv,zhang2023h2o}. % TODO
Although these methods effectively shorten input text to enhance efficiency, they do so at the cost of diminished context integrity, with the loss becoming more pronounced as input texts lengthen.

\paragraph{Context Distillation}

% 1. on short text  \cite{snell2022learning,askell2021general,padmanabhan2024propagating},
% 2. on specific types of context
% 3. use additional parameters and pretraining

Some existing works have explored injecting knowledge in prompt to model parameters.
However, many of these focus on short contexts \cite{snell2022learning,askell2021general,padmanabhan2024propagating}.
Others tend to concentrate on specific types of context such as conversation history \cite{magister2024way}, sports documents \cite{mecklenburg2024injecting}, and task instructions \cite{choi2022prompt}, or they require additional parameters and depend on specialized pretraining \cite{muhtar2024streamadapter,chen2024generative}, which limits their applicability. Additionally, Temp-LoRA \cite{wang2024greater} addresses long-form text generation by storing context information through training the model on the prompt and previously generated content.
In contrast, we aim to present a general approach that is applicable to a wide range of scenarios. Our evaluation covers a variety of tasks, especially involving long contexts, to rigorously assess the effectiveness of context distillation.

\section{Problem Definition}
\subsection{Overview}
Human cognition processes information through two complementary memory systems: short-term memory for transient signal maintenance and long-term memory for persistent knowledge storage. Inspired by this neurocognitive architecture, we propose a novel perspective for language model memory system coordination, with a particular focus on transforming temporary context knowledge into permanent parameter updates, a process that emulates hippocampal-cortical memory consolidation mechanisms \cite{squire1992memory}.

\subsection{Short-term Memory}
\label{short}
Input context in language models parallels with human short-term memory, exhibiting two neurocognitively grounded limitations:
(\textit{i}) Temporal decay: Mirroring the rapid fading of human temporal memory without rehearsal \cite{cowan2008differences}, language models show severe performance degradation as sequence length increases. (\textit{ii}) Capacity bottleneck: Memory constraints enforce strict token limits, analogous to Miller's "magical number seven" in human cognition \cite{miller1956magical}. 

\subsection{Long-term Memory}
\label{long}
Parameters in language models functions as a comprehensive knowledge repository, akin to the vast store of knowledge in human long-term memory. Unlike short-term memory, parameters remains unchanged unless deliberately updated and imposes no additional hardware burdens after fine-tuning.
\subsection{Memory Transformation}
\label{short-to-long}
%Keep expanding the context window is limited by (1) the computational cost of processing long sequences, and (2) the difficulty for current attention mechanisms to retrieve and integrate information across different positions throughout the context.
% in human cognition, short-term memory and long-term memory have different characteristics and functions. Short-term memory is mainly responsible for storing recent events (with maximal details), while long-term memory can store more distant information (with minimal details). This transformation between short-term and long-term memory may be similar to the relationship between context and parameter in the model. When we continuously transform information from contexts into parameters, some memories are spontaneously integrated, some are strengthened, and some gradually fade away.
The process of transforming context into parameters bears a striking resemblance to hippocampal-cortical consolidation observed in human memory systems \cite{squire1992memory}. In this biological process, transient hippocampal memory traces are gradually integrated into neocortical networks through sleep-like replay mechanisms \cite{klinzing2019mechanisms}. During consolidation, neuronal connections undergo dynamic restructuring: new synapses form, while others are pruned. Concurrently, some memories are integrated into existing knowledge structures, others are reinforced, and some gradually attenuate. This parallel between neural network adaptation and biological memory consolidation offers intriguing insights into the nature of learning and memory in both artificial and natural systems.

This transformation offers three key practical advantages: 
(\textit{i}) Efficiency: Converting context into parameters greatly reduces GPU memory usage for KV cache and computational costs for attention.
(\textit{ii}) Persistency: It potentially maintains performance across varying input lengths by avoiding the performance degradation of long-range attention.
(\textit{iii}) Scalability: It theoretically supports infinite context by continuously integrating context information into long-term memory.

Formally, let $P_\theta(y|c)$ denote the distribution generated by the teacher model conditioned on context $c$. We aim to learn parameters $\theta'$ such that the student model's unconditional distribution matches the teacher's conditional distribution:
%$P_{\theta'}(y) = P_{\theta}(y|c) \quad \forall y$.
\begin{equation*}
    P_{\theta'}(y) = P_{\theta}(y|c) \quad \forall y
\end{equation*}
This ensures that the student model internalizes the context $c$ without requiring explicit attention to it during inference. Another perspective is to view the whole parameter set of a Transformer as the state of a meta recurrent neural network (RNN).

\section{Our Framework}
\subsection{Overview}
To tackle the above problem, our framework transforms short-term memory (context) into long-term memory (parameters) through three coordinated phases: 
(\textit{i}) Context knowledge elicitation extracts task-specific knowledge while preserving general sequence completion capabilities via transfer set $\mathcal{T}$ construction.  
(\textit{ii}) Path selection refines $\mathcal{T}$ into a high-quality subset $\mathcal{T}_k$ that maximize knowledge transfer potential through perplexity-guided filtering.  
(\textit{iii}) Memory consolidation aligns the student model’s distribution with the teacher’s context-conditioned outputs over $\mathcal{T}_k$.

\subsection{Context Knowledge Elicitation}
\label{section:context_knowledge_elicitation}
Effective memory transformation necessitates comprehensive contextual understanding beyond superficial pattern extraction. Existing methods that rely on task-specific strategies (e.g., question generation \cite{mecklenburg2024injecting,magister2024way} or knowledge propagation \cite{padmanabhan2024propagating}) are confined to either surface-level patterns or partial facets of contextual knowledge, inherently failing to generalize across tasks and domains. To address this, we construct an elicitation framework designed to systematically excavate multidimensional contextual knowledge through dual objectives:  
(\textit{i}) Diversity: Generate continuations that explore distinct facets of the context (\textit{e.g.}, question answering, summarization).  
(\textit{ii}) Skill specificity: Ensure continuations elicit targeted capabilities (\textit{e.g.}, multi-hop inference, counterfactual reasoning).  

Our approach begins by prompting the model to automatically infer potential queries based on the input context, followed by instructing it to produce context-grounded responsess to these queries.
This process leads to the development of a transfer set $\mathcal{T}$, defined as:
% Formally, the transfer set $\mathcal{T}$ is constructed as:
\begin{equation*}
    \mathcal{T} = \left\{ \left( c, \{q_i\}_{i=1}^n, \{r_i\}_{i=1}^n \right) \middle| \begin{aligned} 
&q_i \sim M_\theta(\cdot|p(c)) \\
&r_i \sim M_\theta(\cdot|c, q_i) 
\end{aligned} \right\}
\end{equation*}
where $c$ represents the input context, $\{q_i\}_{i=1}^n$ is a set of generated queries, and $\{r_i\}_{i=1}^n$ contains the corresponding teacher responses. 
The structured prompt $p(c)$ (Appendix~\ref{appx:implementation}) is meticulously crafted to include (i) a guideline directing the model to anticipate all potential real-user queries concerning the context, (ii) a hinting list of typical types of real-world user queries, (iii) specific examples for each query type, and (iv) output format requirements to facilitate parsing.
% We include possible user query types both before and after the context, as we found that extremely long contexts may cause the model to forget the objective. 

\subsection{Path Selection}
\label{section:path_selection}
Our path selection mechanism promotes effective knowledge transfer by advocating the distributional separation between teacher and student outputs. This approach preserves high-quality continuations that inherently contain richer transferable knowledge while filtering out unreliable samples caused by stochastic generation. Specifically, we implement a divergence-based selection strategy inspired by RHO-1 \cite{lin2024not}.
Formally, for each query-response pair $(q,r)$ generated under context $c$, we compute its perplexity discrepancy as:
\begin{equation*}
\Delta_{\text{PPL}}(r) = \log P_{\theta}(r|c,q) - \log P_{\theta'}(r|q)
\end{equation*}
where $P_{\theta}(\cdot)$ and $P_{\theta'}(\cdot)$ denote the sequence probabilities from the teacher and student models, respectively. We then select the top-$k$ pairs per context that maximize $\Delta_{\text{PPL}}$, creating a refined subset $\mathcal{T}_k$.

\subsection{Memory Consolidation}
This process aims to integrate context knowledge into model parameter updates by reducing the inconsistency between the teacher’s context-aware outputs and the student’s context-agnostic predictions.
Specifically, we optimize the student model $M_{\theta'}$ through knowledge distillation to align with the teacher's context-conditioned probability distribution over $\mathcal{T}_k$:
\begin{equation}
\begin{split}
\mathcal{L} = \mathbb{E}_{(c,Q,R) \sim \mathcal{T}_k} \mathbb{E}_{(q,r) \sim (Q,R)} \mathcal{D}(c, q, r) \\
\mathcal{D}(c, q, r) = f\left( M_{\theta'}(r|q), M_{\theta}(r|c,q) \right)
\end{split}
\label{equation:training}
\end{equation}
where $M(\cdot)$ denotes the model's outputs (\textit{e.g.}, next-token distributions), while $f$ measures the prediction mismatch between teacher and student. We explore three alignment approaches for implementing $f$:  
(\textit{i}) Hidden-state proximity (\textit{e.g.}, mean squared error on hidden states),  
(\textit{ii})) Logit-level distribution matching (\textit{e.g.}, forward/reverse KL divergence), and  
(\textit{iii}) Sequence-level imitation (\textit{e.g.}, fine-tuning on teacher-generated sequences).
Our comprehensive analysis in Section~\ref{ablation_section_loss} reveals task-specific trade-offs between knowledge retention and generalization for these strategies.

\section{Evaluation Setup}
To validate the effectiveness of our approach, we perform controlled experiments under two circumstances: single transformation and sequential transformations. 
While the former isolates the efficacy of individual memory transformation, the latter assesses operational viability under real-world streaming contexts where inputs are gradually received and processed.

\subsection{Single Transformation}
To comprehensively evaluate memory transformation capabilities, we design four task types under this setting, covering diverse real-world scenarios.
The context for all test cases is limited to within $8K$ to meet the context window size restrictions of most LLMs.

\paragraph{Document-based QA}
We use the Natural Questions (NQ) dataset \cite{kwiatkowski2019natural}, which consists of real user queries paired with Wikipedia passages. This task evaluates the model's ability to identify and retain factual knowledge from long-form documents. We treat the Wikipedia passages as the contexts.

\paragraph{Many-shot In-context Learning}
To evaluate the model's skill acquisition capabilities, we construct many-shot test datasets from three classification tasks: TREC Coarse\footnotemark[1]\footnotetext[1]{\url{https://huggingface.co/datasets/CogComp/trec}} ($6$-class question categorization), TREC Fine\footnotemark[1] ($50$-class question typing), and NLU\footnotemark[2]\footnotetext[2]{\url{https://huggingface.co/datasets/xingkunliuxtracta/nlu_evaluation_data}} ($68$-class intent detection). To prevent information leakage from pre-training data, we replace original labels with randomly assigned numerical values. We randomly sample 300 examples from the training set of each dataset (i.e., 300-shots learning) as the contexts.

\paragraph{Knowledge Update.}
We evaluate the model's grounded reasoning capabilities using counterfactual knowledge updates. Two datasets are used: (\textit{i}) CounterFact \cite{meng2022locating}, which contains factual statements that contradict established knowledge, and (\textit{ii}) MQuAKE \cite{zhong2023mquake}, which focuses on multi-hop question answering under sequential knowledge updates. We treat the factual statements in CounterFact and the concatenated sentences of updates in MQuAKE as the contexts.

\paragraph{Text Generation}
We evaluate text generation capabilities using the test set of PG19 book corpus \cite{rae2019compressive}. For each test case, we extract a $2048$-token segment as context and the subsequent $256$-token segment as the target continuation. 

\begin{table*}[t]
\small
\centering
\setlength{\tabcolsep}{2pt}
\begin{tabular}{ccccccccc}
\toprule
\multirow{2}{*}{\textbf{Method}} & \textbf{Doc-based QA} & \multicolumn{3}{c}{\textbf{Many-shot ICL}} & \multicolumn{2}{c}{\textbf{Knowledge Update}} & \multirow{2}{*}{\textbf{Avg.}} & \textbf{Text Generation} \\
& \textbf{NQ} & \textbf{trec\_fine} & \textbf{trec\_coarse} & \textbf{nlu} & \textbf{counterfact} & \textbf{mquake} & & \textbf{PG19} \\
\midrule
\rowcolor{gray!20}
\multicolumn{9}{c}{\textbf{\textit{Upper Bound}}} \\
\midrule
Full Context & \underline{53.6(1)} & \underline{61.2(1)} & 79.2(1) & \underline{78.4(1)} & 46.8(1) & 92.4(1) & 68.6(1) & \underline{14.6(1)} \\
\midrule
\rowcolor{gray!20}
\multicolumn{9}{c}{\textbf{\textit{Context Compression}}} \\
\midrule
Local Context & 31.4(.31) & 36.6(.60) & 65.2(.82) & 33.0(.42) & 17.5(.37) & 24.8(.18) & 34.8(.47) & 19.9(.34) \\
Selective Context & 20.2(-.04) & 0.4(.00) & 0.0(0) & 3.0(.04) & 6.8(.14) & 10.8(.01) & 6.9(.03) & 326.3(-38.36) \\
LLMLingua & 19.6(-.06) & 2.8(.04) & 23.0(.29) & 15.7(.20) & 47.1(1.01) & 92.0(1.00) & 33.4(.44) & 70.7(-6.08) \\
LLMLingua-2 & 29.6(.25) & 0.4(.00) & 0.4(.01) & 0.3(.00) & 6.6(.14) & 25.0(.19) & 10.4(.08) & 57.7(-4.44) \\
SnapKV & 0.4(-.65) & 0.8(.01) & 20.4(.26) & 0.2(.00) & 1.2(.02) & 21.0(.14) & 7.3(.03) & 625.9(-76.19) \\
% SnapKV (query-informed) & 51.4(.93) & 57.8(.94) & 76.6(.97) & 73.9(.94) & 46.4(.99) & 15.6(.07) & 53.6(.76) & 14.6(1) \\
ReadAgent & 43.0(.67) & 17.2(.28) & 16.0(.20) & 5.5(.07) & 0.2(-.00) & 11.2(.02) & 15.5(.16) & 20.8(.22) \\
\midrule
\rowcolor{gray!20}
\multicolumn{9}{c}{\textbf{\textit{Context Distillation}}} \\
\midrule
Temp-LoRA & 37.2(.49) & 12.4(.20) & 20.8(.26) & 4.2(.05) & 39.6(.85) & 1.6(-.10) & 19.3(.22) & 19.9(.34) \\
Ours & \textbf{45.2(.74)} & \textbf{59.0(.96)} & \textbf{83.4(1.05)} & \textbf{72.7(.93)} & \textbf{66.9(1.43)} & \textbf{95.4(1.04)} & \textbf{70.4(1.03)} & \textbf{17.9(.59)} \\
\midrule
\rowcolor{gray!20}
\multicolumn{9}{c}{\textbf{\textit{Lower Bound}}} \\
\midrule
No Context & 21.4(0) & 0.2(0) & 0.0(0) & 0.1(0) & 0.3(0) & 9.6(0) & 5.3(0) & 22.6(0) \\
\bottomrule
\end{tabular}
\caption{Results for single transformation tasks: PPL is reported for text generation and EM for other tasks. Each entry $x(y)$ indicates $x$ as the model’s performance ($\times 100\%$) and $y$ as the recovery rate relative to the bounds. Underlined values indicate unsurpassed full-context prompting, and bold indicates the best among other methods.}
\label{main_results}
% \vspace{-2mm}
\end{table*}
% TODO: 8K context.

\subsection{Sequential Transformations}
In this setting, we aim to test memory transformation under realistic scenarios that involve: (\textit{i}) Long-context processing: Inputs exceeding the context window, requiring multiple transformations. (\textit{ii}) Real-world complexity: Diverse task types and input lengths, covering a range of difficulties. To meet these criteria, we adopt LongBench v2 \cite{bai2024longbench}, a comprehensive benchmark for long-context understanding. LongBench v2 spans diverse task types, including QA, ICL, dialogue understanding, code repository understanding, and structured data understanding, curated from real-world domains such as legal documents, scientific articles, and technical manuals. Context lengths range from $8$K to $2$M words, challenging models to handle extreme input sizes.

\section{Experiments}

\paragraph{Implementation Details} 
We employ Llama3-8B-instruct ($8$K context window) as our base model. The standard implementation of our method follows this protocol: (\textit{i}) Generate $200$ query-response pairs and $200$ open-ended continuations for each context (Section \ref{section:context_knowledge_elicitation}), with each entry consisting of 512 tokens. (\textit{ii}) Retain the top $200$ entries as $\mathcal{T}_k$ (Section \ref{section:path_selection}). 
(\textit{iii}) Transform the context into parameter updates via forward KL divergence.
To establish a standardized evaluation protocol across all context compression and distillation methods, we define retention ratio as $\rho = \frac{|c'|}{|c|}$, representing the proportion of retained content $c'$ from the original context $c$. Unless otherwise specified, we set $\rho=0.1$ in our experiments. For context compression baselines, $c'$ is determined by the model, while for context distillation baselines and our method, $c'$ comprises the last $\rho$ proportion of the context.
Further details are provided in Appendix \ref{appx:implementation}.

\subsection{Baselines}
\paragraph{Upper Bound (Full Context)}
Full-context prompting serves as the theoretical upper bound, ensuring complete contextual integrity. For the LongBench v2 data, where the context may exceed the context window size, we additionally use Llama3.1-8B-instruct (extended to $128$K context) with middle-truncation \citet{bai2024longbench} as a practical upper bound.

\paragraph{Lower Bound (No Context)}
Context-agnostic performance (no context provided) serves as the universal lower bound across all tasks.

\paragraph{Context Compression} ReadAgent \cite{lee2024human} segments and summarizes input, while Selective Context \cite{li2023compressing}, LLMLingua \cite{jiang2023llmlingua}, and LLMLingua-2 \cite{pan2024llmlingua} directly eliminate redundant tokens. SnapKV \cite{li2024snapkv} retains the most important key-value pairs within the context's KV cache.
We also incorporate a practical baseline called Local Context, which retains the last $\rho$ proportion of the context, allowing us to isolate the benefits of knowledge transformation from context retention effects.

\paragraph{Context Distillation} We employ Temp-LoRA \cite{wang2024greater}, a method that directly fine-tunes the model on the context.

\subsection{Metrics}  
We evaluate performance using task-specific metrics: exact match (EM) for document-based QA, many-shot ICL, and knowledge update tasks, perplexity (PPL) for text generation, and accuracy for Longbench v2. We exclude text generation when reporting average single transformation performance due to differing metric scales. To quantify the effectiveness of memory transformation, we introduce the recovery rate, defined as:
\begin{equation*}
R = \frac{M - L}{U - L} %\times 100\%
\end{equation*}
where $M$, $U$, and $L$ denote method performance, upper bound, and lower bound respectively. This unitless metric enables cross-task comparison of context utilization efficiency.

\subsection{Single Transformation Tasks}
We systematically evaluate memory transformation efficacy and assess individual component impacts within the controlled single-transformation setting.

\begin{figure}[t]
 \centering
 \includegraphics[width=\linewidth]{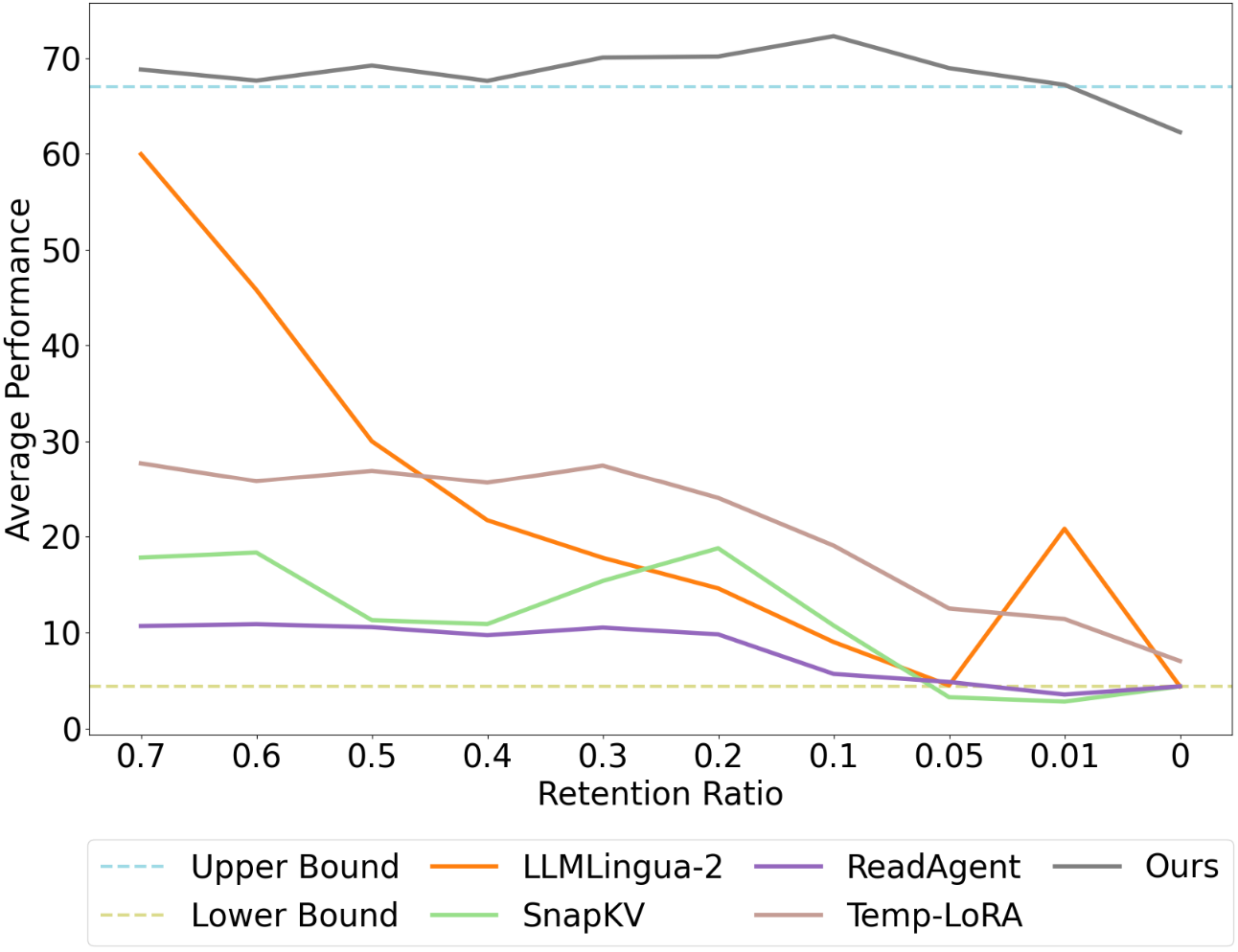}
 \caption{
Average model performance on single transformation tasks across different retention ratios. To maintain the readability of the figure, we only show the results of some representative baselines.
}
 \label{fig:transform_ratio}
% \vspace{-2mm}
\end{figure}

\paragraph{Results}
The results in Table \ref{main_results} reveal three key findings: (\textit{i}) First, while full-context prompting establishes an upper-bound performance benchmark by enabling full-context attention, our method demonstrates unexpected superiority in specific tasks, achieving an average performance recovery rate of $103\%$. Notably, in reasoning-intensive scenarios such as many-shot ICL ($98\%$ average recovery) and knowledge updates ($123\%$ average recovery), our approach shows remarkable effectiveness.
(\textit{ii}) Second, comparison with the Local Context baseline confirms that the performance gains of our method arise from permanent parameter updates rather than simply leveraging recent context segments. This distinction is important, as several alternative baselines perform poorly, with some even failing to surpass the lower bound.
(\textit{iii}) The third finding highlights domain-specific limitations: in document-based QA (NQ) focused on fact recall, while outperforming all baselines, our method achieves only $74\%$ performance recovery. This aligns with human cognitive processing, where converting context into knowledge prioritizes comprehension over rote memorization, especially when queries are underspecified. 

\paragraph{Performance vs. Retention Ratio}
Building on initial validation, we further analyze how retention ratio affects model performance to identify operational boundaries. We vary the ratio from $0.7$ to $0$, representing the gradual compression/parameterization of the context.
Figure \ref{fig:transform_ratio} shows our method maintains stable performance across ratios ($68.8\rightarrow62.3$; peak $72.3$ at a $0.1$ ratio), while all baseline methods degrade and collapse at extreme ratios. This also evidences our approach's capacity to leverage transformed long-term memory instead of relying on residual context fragments.

\begin{table}[t]
\centering
\small
\setlength{\tabcolsep}{13pt}
\begin{tabular}{c|ccc}
\toprule
Method & N=50 & N=100 & N=200 \\
\midrule
{Random} & 68.54 & 69.83 & 69.91 \\
{KL-based} & 66.54 & 68.12 & 70.60 \\
{PPL-based} & \textbf{69.61} & \textbf{70.47} & \textbf{72.27} \\
\bottomrule
\end{tabular}
\caption{Ablation of path selection methods, with models trained on $N$ continuations retained from $400$.}
\label{path_selection}
% \vspace{-2mm}
\end{table}

\paragraph{Effect of path selection strategies}
\label{ablation_section_loss}
To show the effect of our PPL-based path selection strategy, we perform an ablation study using $10\%$ of the data from each task.
Results in Table~\ref{path_selection} highlight our method's superiority over both random selection and KL divergence-based choices in memory transformation. Our PPL-differential strategy consistently outperforms (peak $72.27$ at $N=200$), surpassing both random selection ($+2.36$ points) and the KL-based method ($+1.67$ points) at $N=200$. Notably, the PPL-based selection exhibits a progressive quality improvement ($69.61\rightarrow72.27$), suggesting its effectiveness in identifying high-information-density continuations. 
We also investigate the impact of retaining different numbers and lengths of continuations on model performance and find that selecting top-$200$ continuations, each $512$ tokens long, yields the best results. The results are summarized in Appendix \ref{appx:continuation_num_length}.

\begin{table}[t]
\small
\centering
\setlength{\tabcolsep}{5.8pt}
\begin{tabular}{cccccc}
\toprule
\textbf{Method} & \textbf{QA} & \textbf{ICL} & \textbf{KU} & \textbf{Avg.} & \textbf{Text Gen} \\
\midrule
\rowcolor{gray!20}
\multicolumn{6}{c}{\textbf{\textit{Upper Bound}}} \\
\midrule
Full Context & \underline{54.0} & 72.9 & 64.5 & 67.0 & 18.2 \\
\midrule
\rowcolor{gray!20}
\multicolumn{6}{c}{\textbf{\textit{Logit-level Alignment }}} \\
\midrule
Forward KL & 46.0 & \textbf{73.5} & \textbf{83.5} & \textbf{72.3} & 22.7 \\
Reverse KL & 44.0 & 72.5 & 80.0 & 70.3 & 23.5 \\
Adaptive KL & 44.0 & 72.7 & 83.0 & 71.4 & 23.0 \\
DPKD & 38.0 & 58.9 & 64.0 & 57.1 & 342.3 \\
\midrule
\rowcolor{gray!20}
\multicolumn{6}{c}{\textbf{\textit{Hidden-state-level Alignment}}} \\
\midrule
MSE & 26.0 & 63.3 & 77.5 & 61.8 & 23.8 \\
\midrule
\rowcolor{gray!20}
\multicolumn{6}{c}{\textbf{\textit{Sequence-level Imitation}}} \\
\midrule
SeqKD & \textbf{48.0} & 68.9 & 35.5 & 54.3 & \textbf{5.5} \\
\midrule
\rowcolor{gray!20}
\multicolumn{6}{c}{\textbf{\textit{Lower Bound}}} \\
\midrule
No Context & 20.0 & 0.1 & 3 & 4.4 & 27.6 \\
\bottomrule
\end{tabular}
\caption{Ablation of transformation loss. We report the average performance for each single transformation task across its datasets, with task names abbreviated.}
\label{ablation_loss}
% \vspace{-2mm}
\end{table}

\paragraph{Impact of the transformation loss}
We examine six transformation loss variants for our framework, using the same aforementioned $10\%$ of data from each task for experimentation. Detailed descriptions of these loss functions are provided in Appendix \ref{appx:loss_variants}, with the results summarized in Table \ref{ablation_loss}.
Our key findings reveal task-dependent trade-offs:
(\textit{i}) Hidden-state-level alignment is inadequate for fact recall: Results on NQ demonstrate that logit-level distribution matching and sequence-level imitation better preserve factual relationships than hidden-state proximity, \textit{e.g.}, FKL outperforms MSE by $20\%$.
(\textit{ii}) Sequence-Level imitation fails in reasoning: SeqKD collapses catastrophically in multi-hop reasoning ($6.0$ vs. FKL’s $96.0$ on Mquake), validating that per-token alignment better preserves causal dependencies.  
(\textit{iii}) PPL’s narrow scope: While SeqKD excels in text generation ($5.5$ PPL vs. FKL’s $22.7$ on PG19), this metric fails to capture holistic performance, especially in compositional reasoning. 
Notably, FKL and MSE display minimal differences on PG19 ($22.7$ vs. $23.8$), yet diverge significantly in other tasks, underscoring the need for multi-dimensional evaluation.
These results establish FKL as the most balanced choice.
% , achieving $108\%$ recovery rate in average performance.

\begin{table*}[t]
\small
\centering
\setlength{\tabcolsep}{12pt}
\begin{tabular}{ccccccc}
\toprule
\multirow{2}{*}{\textbf{Method}} & \multirow{2}{*}{\textbf{Overall}} & \multicolumn{2}{c}{\textbf{Difficulty}} & \multicolumn{3}{c}{\textbf{Length (<32K; 32K-128K; >128K)}} \\
&  & \textbf{Easy} & \textbf{Hard} & \textbf{Short} & \textbf{Medium} & \textbf{Long} \\
\midrule
\rowcolor{gray!20}
\multicolumn{7}{c}{\textbf{\textit{Upper Bound}}} \\
\midrule
128K Context$^{\clubsuit}$ & 30.0(1) & 30.7(1) & 29.6(1) & 35.0(1) & 27.9(1) & 25.9(1) \\
\midrule
\rowcolor{gray!20}
\multicolumn{7}{c}{\textbf{\textit{Context Compression}}} \\
\midrule
8K Context & 18.5(-.57) & 21.9(.00) & 16.4(-1.05) & 17.2(-1.47) & 20.0(0) & 17.6(-.28) \\
Local Context & 15.5(-.97) & 15.6(-.71) & 15.4(-1.20) & 22.2(-.77) & 10.7(-1.18) & 13.9(-.86) \\
Selective Context & 16.3(-.87) & 16.2(-.65) & 16.4(-1.05) & 21.1(-.92) & 13.5(-.82) & 13.9(-.86) \\
LLMLingua & 22.7(0) & 25.0(.35) & 21.2(-.30) & 26.7(-.15) & 21.9(.24) & 17.6(-.29) \\
LLMLingua-2 & 22.3(-.05) & 21.9(0) & 22.5(-.10) & 28.9(.15) & 16.7(-.41) & 22.2(.43) \\
SnapKV & 1.4(-2.90) & 1.7(-2.86) & 0.0(-3.09) & 1.0(-2.95) & 1.9(-2.83) & 1.6(-2.87) \\
ReadAgent & 16.1(-.89) & 16.2(-.65) & 16.1(-1.10) & 21.7(-.85) & 14.9(-.65) & 9.3(-1.58) \\
\midrule
\rowcolor{gray!20}
\multicolumn{7}{c}{\textbf{\textit{Context Distillation}}} \\
\midrule
Temp-LoRA & 32.4(1.33) & 34.4(1.42) & 31.2(1.25) & 33.3(.77) & 27.0(.88) & 41.7(3.44) \\
Ours & \textbf{44.7(3.01)} & \textbf{49.5(3.13)} & \textbf{41.8(2.89)} & \textbf{41.7(1.92)} & \textbf{46.1(3.30)} & \textbf{47.2(4.30)} \\
\midrule
\rowcolor{gray!20}
\multicolumn{7}{c}{\textbf{\textit{Lower Bound}}} \\
\midrule
No Context & 22.7(0) & 21.9(0) & 23.2(0) & 27.8(0) & 20.0(0) & 19.4(0) \\
\bottomrule
\end{tabular}
\caption{Results for sequential transformation tasks: We report accuracy, as all data is organized in a multiple-choice QA format. Each entry $x(y)$ indicates $x$ as the model’s performance ($\times 100\%$) and $y$ as the recovery rate relative to the bounds. ${\clubsuit}$: results from \citet{bai2024longbench}.}
\label{longbench_results}
% \vspace{-2mm}
\end{table*}

\subsection{Sequential Transformations Tasks}
We extend our evaluation to complex, real-world scenarios to gain insights into practical contexts.
\paragraph{Results}
As shown in Table \ref{longbench_results}, our method consistently outperforms all other approaches across various categories, including overall performance, difficulty levels, and text lengths. Notably, our method demonstrates remarkable efficacy in handling longer contexts, as evidenced by the substantial performance gains in the "Medium" and "Long" categories. This suggests a robust capacity for managing extended sequences, traditionally challenging due to memory constraints. Furthermore, the substantial improvements observed in both the "Easy" and "Hard" difficulty categories indicate the model's adaptability and effectiveness across varying task complexities.
In comparison to $128$K Context and context compression methods, the significant performance advantages achieved by our method underscore the efficacy of memory transformation in enhancing contextual understanding and application. 

\begin{figure}[t]
 \centering
 \includegraphics[width=\linewidth]{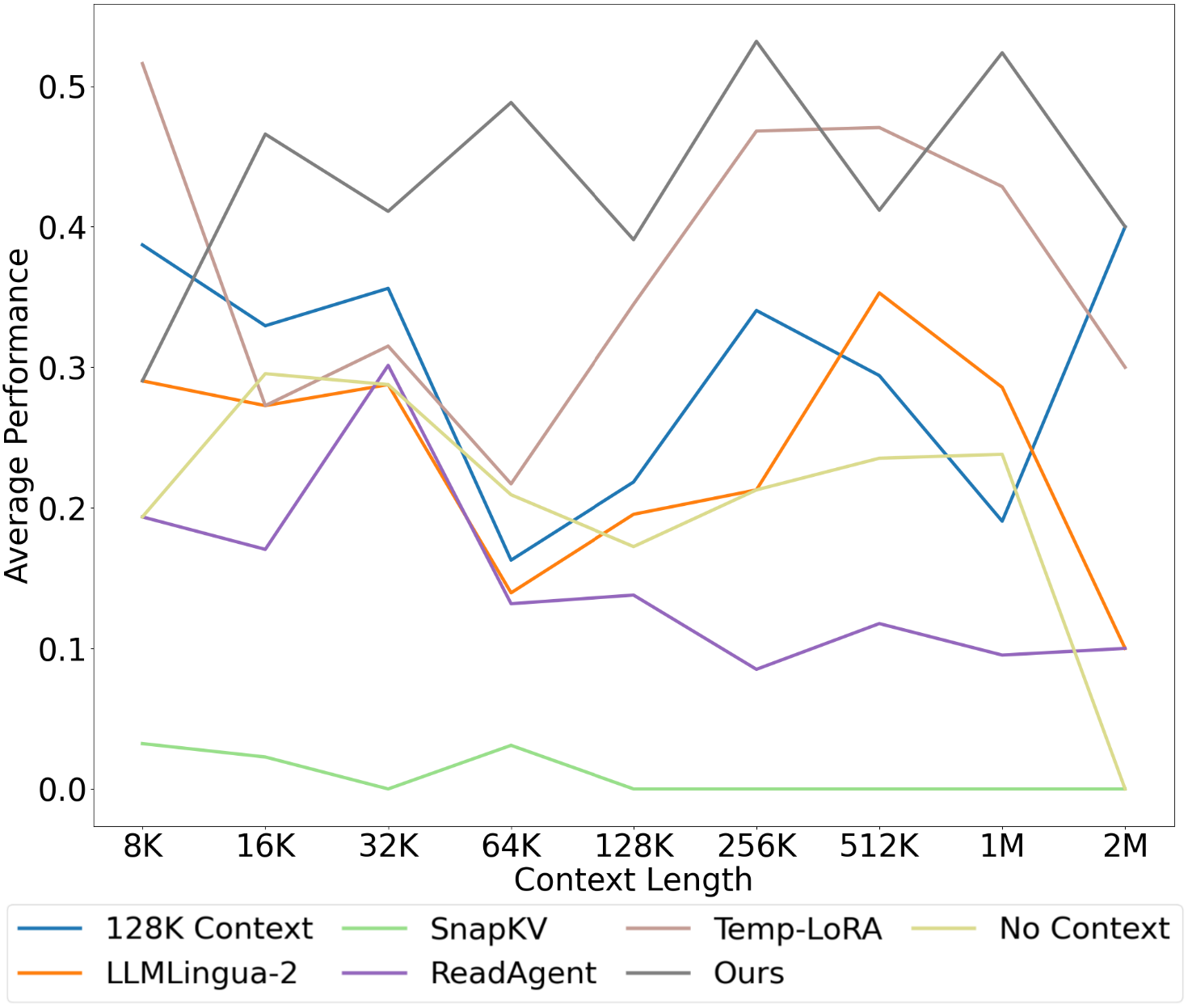}
 \caption{Performance comparison across different context lengths in sequential transformation tasks. To maintain the readability of the figure, we only show the results of some representative baselines.
}
\label{fig:context_length}
% \vspace{-2mm}
\end{figure}

\paragraph{Performance vs. Context Length}
Given that contexts of varying lengths can present different challenges, we analyze the stability and scalability of all methods in processing these variations.
As shown in Figure \ref{fig:context_length}, where LongBench v2 results are reorganized by context length, baseline methods exhibit severe performance degradation (average decline: $15.4\% \pm 12.8\%$) within the $8$K–$64$K range, while our method demonstrates markedly greater stability, even as contexts extend to $2$M. Notably, a paradoxical performance spike occurs in certain baselines at the $512$K length. We attribute this anomaly to pretrained priors rather than genuine context utilization capabilities, as evidenced by analogous improvements in no-context performance at the same length. When contexts further scale to $1$M–$2$M, baselines suffer catastrophic failure, whereas our method sustains robust performance. These findings underscore our method’s architectural advances in persistently handling contexts far exceeding conventional window sizes.

\begin{table}[t]
\centering
\small
\setlength{\tabcolsep}{7pt}
\begin{tabular}{ccc}
\toprule
\textbf{Length per Turn} & \textbf{Avg Turn} & \textbf{Overall Performance} \\
\midrule
1k & 277 & 45.54\\
2k & 138 & 43.75 \\
3k & 93 & 44.64 \\
4k & 69 & 46.43 \\
5k & 53 & 49.11 \\
6k & 46 & 46.43 \\
\bottomrule
\end{tabular}
\caption{Average transformation turns and performance by context length per turn.}
\label{chunk_size}
% \vspace{-2mm}
\end{table}

\paragraph{Impact of Context Length per Transformation Turn}
To analyze how context segmentation granularity affects sequential transformation efficacy, we uniformly sample $100$ contexts ($112$ associated questions) from Longbench v2. As shown in Table~\ref{chunk_size}, performance peaks at $49.11$ when processing $5$K-token chunks, requiring fewer transformation turns (Avg$=53$). Shorter chunks degrade performance due to fragmented knowledge integration, while longer chunks ($6$K) reduce performance by $2.68$ points, likely due to information overload affecting the model's capacity to consolidate knowledge. These results suggest an optimal range of around $5$K tokens for an $8$K context window model, effectively balancing semantic coherence and manageable information load.

\section{Conclusion}
This work parallels LLMs' context and parameters with human memory systems, establishing a framework to transform temporary context knowledge into permanent parameter updates through elicitation, path selection, and memory consolidation.
In single transformation tasks, our method remarkably surpasses full-context prompting performance, achieving an average $103\%$ recovery rate. Further analysis reveals that baselines exhibit brittle performance under high compression ratios, whereas our method maintains robust performance, even maintaining $62.25$ performance at extreme ratios where most baselines collapse to no-context performance.
Our framework's generalizability is further evidenced by its consistent performance across varying context lengths and task difficulties in sequential transformation tasks, highlighting its applicability in real-world scenarios.
% This underscores the method's potential to enhance the scalability and efficiency of NLP systems by effectively managing extensive contexts.

\section*{Limitations}  
While our framework demonstrates significant advantages in memory transformation, several limitations warrant discussion:  
\paragraph{Model Diversity} Our experiments are conducted primarily using Llama-3-8B-instruct, which may limit the generalizability of our findings. A broader range of models could provide additional insights. 
\paragraph{Compute Cost} Our framework currently encounters limitations regarding computational efficiency during gradient-based fine-tuning. Future work may consider to use hypernetwork for much efficient transformation.

\bibliography{custom}

\appendix

\section{Appendix}
\label{sec:appendix}

\subsection{Implementation Details}
\label{appx:implementation}
During the context knowledge elicitation phase, we employ the following hybrid prompt to guide the model in generating high-quality queries:

\begin{promptbox}
Please prepare to analyze the text provided below. As you read, simulate real-world user queries about the content, such as summarizing, detailing, or inferring knowledge. \\
For example, consider the following potential user queries if applicable to the provided text: \\
1. Ask for a concise summary that captures the main points and essential details of the text. Anticipate user requests such as, "What are the central arguments?" or "Can you summarize the main events of the story?" \\
2. Formulate questions about key details or themes within the text, such as "What achievements did Anthony Joshua achieve in boxing?" or "What was the date and venue of event?" \\
3. Identify and explore patterns or examples, create similar formatted examples or pose questions, such as "What is the labeling criteria for these examples?" \\
4. Integrate and reflect on new knowledge from the text, asking for the implications and applications of the new knowledge. \\
5. Request repetition of specific sentences or paragraphs from the text. \\
6. ... \\
Here is the text for analysis: \\
======= Text Begins Here ======= \\
\{context\} \\
======= Text Ends Here ======= \\
Let's review the potential user queries to see if they apply to the provided text. \\
1. Ask for a concise summary that captures the main points and essential details of the text. Anticipate user requests such as, "What are the central arguments?" or "Can you summarize the main events of the story?" \\
2. Formulate questions about key details or themes within the text, such as "What achievements did Anthony Joshua achieve in boxing?" or "What was the date and venue of event?" \\
3. Identify and explore patterns or examples, create similar formatted examples or pose questions, such as "What is the labeling criteria for these examples?" \\
4. Integrate and reflect on new knowledge from the text, asking for the implications and applications of the new knowledge. \\
5. Request repetition of specific sentences or paragraphs from the text. \\
6. ... \\
Determine suitable query types for the text and generate a comprehensive set of queries that thoroughly cover the content.  \\
Make the subject of the query clear and avoid using pronouns like "it," "he," or "she" to prevent ambiguity. \\
Output 20 queries directly, each on a separate line, numbered from "1." to "20." Conclude with "|||||" as the end symbol.
\end{promptbox}

We set the temperature to $1$ for sampling queries and responses to enhance diversity while maintaining quality.
In the memory consolidation phase, we train the model using LoRA with parameters: lora\_rank$=8$, lora\_alpha$=16$, and lora\_dropout$=0.05$, with a learning rate of $1e-4$. The dataset $\mathcal{T_k}$ is divided into training and development sets at a $4:1$ ratio. After each epoch, we assess the dev set loss, stopping training if the loss does not decrease after two consecutive evaluations. In the sequential transformation setting, each transformation adheres to the same protocol. For testing, we set the temperature to 0 (\textit{i.e.}, greedy decoding) to ensure reproducibility.

\subsection{loss variants}
\label{appx:loss_variants}
(\textit{i}) Forward KL (FKL) minimizes $\text{KL}(M_\theta(y|c,x) \parallel M_{\theta'}(y|x))$ per token, enforcing precise distribution matching.  
(\textit{ii}) Reverse KL (RKL) optimizes $\text{KL}(M_{\theta'}(y|x) \parallel M_\theta(y|c,x))$, prioritizing mode coverage over exact alignment.  
(\textit{iii}) Adaptive KL (AKL) \cite{wu2024rethinking} dynamically blends FKL/RKL weights based on teacher-student distribution divergence.  
(\textit{iv}) DPKD \cite{li2024direct} incorporates preference-driven RKL with length normalization.  
(\textit{v}) MSE aligns hidden states between models, bypassing distributional objectives.  
(\textit{vi}) SeqKD applies sequence-level teacher forcing over sampled continuations.

\begin{table}[t]
\centering
\small
\setlength{\tabcolsep}{10pt}
\begin{tabular}{cc|cccc}
\toprule
\multicolumn{2}{c|}{\diagbox{N}{L}} & 128 & 256 & 384 & 512 \\
\midrule
\multicolumn{2}{c|}{100} &  64.13 & 65.37 & 67.00 & 70.47  \\
\multicolumn{2}{c|}{200} &  66.47 & 66.30 & 69.67 & 72.27 \\
\multicolumn{2}{c|}{300} & 66.30 & 67.10 & 68.17 & 69.53 \\
\multicolumn{2}{c|}{400} & 66.05 &  67.35 &  68.95 &  69.40 \\
\bottomrule
\end{tabular}
\caption{Ablation: Continuation Count and Length. Each set of $N$ continuations comprises an equal mix of query-response pairs and open-ended continuations. }
\label{table:continuation_num_length}
\end{table}

\subsection{Impact of the number and length of the continuations.}
\label{appx:continuation_num_length}
We assess the impact of $\mathcal{T}_k$'s scale on memory consolidation. To ensure that extending count/length settings provides additional information, each setup includes all content from the smaller or shorter settings. The results in Table \ref{table:continuation_num_length} reveal a non-linear relationship between continuation configurations and model performance. Performance generally improves with longer continuations (L), peaking at $72.27$ for $N=200$ and $L=512$. However, a quality-quantity trade-off emerges: the performance inversion between $N=200,L=512$ ($72.27$) and $N=300,L=512$ ($69.53$) suggests excessive continuation counts degrade information density, emphasizing that high-quality continuations outweigh sheer quantity for effective knowledge consolidation.

\end{document}